\documentclass{article}

    \PassOptionsToPackage{numbers, compress}{natbib}

\newcommand{\revision}{\textcolor{black}}

    \usepackage[final]{timeseries_workshop}

\usepackage[utf8]{inputenc} 
\usepackage[T1]{fontenc}    
\usepackage{hyperref}       
\usepackage{url}            
\usepackage{booktabs}       
\usepackage{amsfonts}       
\usepackage{nicefrac}       
\usepackage{microtype}      
\usepackage{xcolor}         
\usepackage{graphicx}
\usepackage{caption}
\usepackage{subcaption}
\usepackage{amsmath}
\usepackage{newfloat}

\DeclareFloatingEnvironment[]{supptable}
\DeclareFloatingEnvironment[]{suppfigure}
\title{Transformer-based Time-Series Biomarker Discovery for COPD Diagnosis}

\author{%
  Soham Gadgil\thanks{Correspondence to <\href{mailto:sgadgil@cs.washington.edu}{sgadgil@cs.washington.edu}>.} \\
  University of Washington\\
  \And
  Joshua Galanter \\
  Genentech, Inc. \\
  \AND
  Mohammadreza Negahdar \\
  Genentech, Inc. \\
}

\begin{document}

\maketitle
 
\vspace{-1.2em}
\begin{abstract}
   Chronic Obstructive Pulmonary Disorder (COPD) is an irreversible and progressive disease which is highly heritable. Clinically, COPD is defined using the summary measures derived from a spirometry test but these are not always adequate. Here we show that using the high-dimensional raw spirogram can provide a richer signal compared to just using the summary measures. We design a transformer-based deep learning technique to process the raw spirogram values along with demographic information and predict clinically-relevant endpoints related to COPD. Our method is able to perform better than prior works while being more computationally efficient. Using the weights learned by the model, we make the framework more interpretable by identifying parts of the spirogram that are important for the model predictions. Pairing up with a board-certified pulmonologist, we also provide clinical insights into the different aspects of the spirogram and show that the explanations obtained from the model align with underlying medical knowledge.
\end{abstract}

\section{Introduction}
Chronic Obstructive Pulmonary Disorder (COPD) is a progressive lung condition characterized by persistent airflow limitation that impairs breathing and is typically caused by long-term exposure to harmful substances like cigarette smoke \cite{ferrera2021advances, macnee2006pathology}. It is the third leading cause of death worldwide, with a global prevalance of 3.91\% (>250 million) \cite{quaderi2018unmet}. Therefore, timely and accurate diagnosis of COPD is crucial to reducing patients' health risks.

Traditionally, the diagnosis of COPD is confirmed using a spirometry test, which measures the flow and volume of air that can be inhaled and exhaled. There are a couple of important summary measures derived from the spirogram: (i) FEV$_1$, or the forced Expiratory Volume in 1 second, which is the volume of air a person can forcefully exhale in the first second of exhalation (ii) FVC, or the forced vital capacity, which is the total amount of air a person can forcefully exhale (iii) PEF, or the peak expiratory flow, which is the maximum flow of air a person can forcefully exhale. A post-bronchodilatory ratio of FEV$_1$ to FVC less that 0.7 is considered to be diagnostic of COPD  in patients with appropriate symptoms and exposures \cite{toren2021ratio, hoesein2011lower}. However, there can be some challenges associated with using such pulmonary function test markers for COPD diagnosis. First, while they identify diagnosed COPD cases, they do not effectively detect patients at an earlier stage or those at risk of developing COPD. Second, the FEV$_1$/FVC ratio is the current gold standard for COPD detection, but it might not always be accurate since there is a lot of heterogeneity involved \cite{bhatt2023fev1, medbo2007lung}. Third, there can be a number of individuals with the pathological hallmarks of COPD who nonetheless fall above the arbitrary threshold, particularly patients with PRISM (preserved ratio impaired spirometry) \cite{higbee2022prevalence}. Fourth, the FEV$_1$/FVC ratio is subject to effort dependence and there can be variability in measurement across different efforts for the same patient. 

Recently, some studies have explored the use of deep learning methodologies to determine whether raw spirogram values can provide a stronger prognostic signal for predicting clinically relevant COPD-related outcomes compared to traditional summary measures \cite{mei2024deep, cosentino2023inference}. In this work, we show that a transformer-based architectural change in one of the prior methods, DeepSpiro \cite{mei2024deep}, performs better across most of the endpoints that we tested while being $\sim$4.5 times more computationally efficient. By using the attention weights of the transformer, we are able to identify parts of the spirometry curve which are important for the different prediction tasks. Leveraging the expertise of a board-certified pulmonologist, we also provide clinical insights into the model predictions and find that the important parts of the curve are confirmed by medical knowledge.

\section{Methods}

\subsection{Dataset and Label Generation}
We use the UK Biobank dataset for this study \cite{biobank2014uk}, which is a large scale biomedical database with health information from around 500K participants. Specifically, we use the spirometry information from UKB field 3066, which includes the expiratory volumes in milliliters recorded every 10 milliseconds. Although the same participant may have multiple visits, we consider the Time-Volume measurements only from their first visit. After multiple preprocessing steps (details in Appendix \ref{preprocess}), we convert the Volume-Time curves to Flow-Volume curves which are the input to our model. The demographic features we use are age, sex, smoking status, and height (collected on the spirometry test date) since they are recognized as being predictive of COPD by our pulmonologist collaborator.

We test three clinical relevant COPD-related endpoints as part of this study: COPD risk, exacerbation, and mortality. COPD risk refers to the likelihood of developing COPD after the spirometry test date. Hospitalization is used as a proxy for exacerbation and mortality indicates a COPD-related death of the patient after the initial visit. To extract these labels from the UK Biobank dataset, we use procedure similar to the one described in \cite{cosentino2023inference}. Specifically, we derive binary COPD labels from the medical records available about the patients, including the primary-care reports, the hospital inpatient records, and the self-report records. Appendix \ref{label_details} provides more details. This study has been conducted using the UK Biobank Resource under Application Number 44257.

\subsection{Time-Series Transformer}
\begin{figure}[h!]
    \centering
    \includegraphics[width=\textwidth]{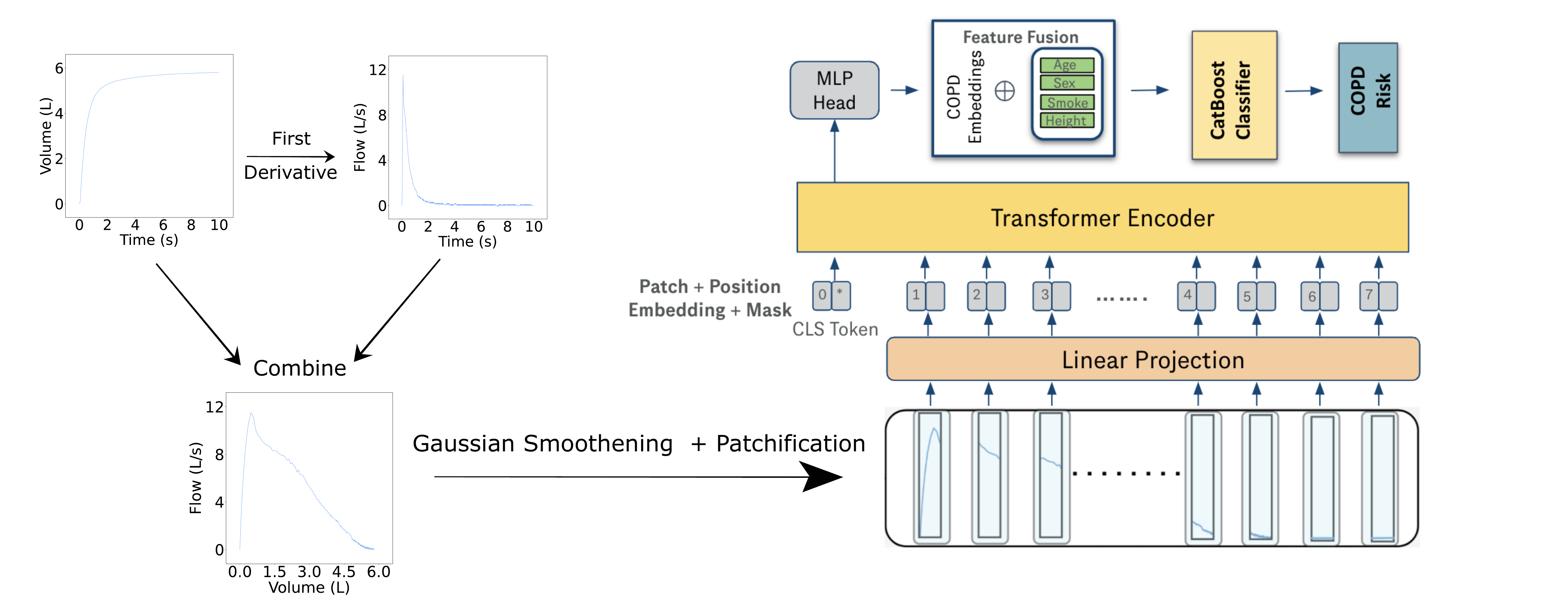}
    \caption{The overall framework that we use. We  start with the raw Volume-Time curve and convert it to a Flow-Volume curve. This curve is then smoothened using a Gaussian filter and patchified before passing as an input to the time-series transformer.}
    \label{fig:arch_fig}
\end{figure}

Here we describe the modeling procedure that we use for the preprocessed dataset obtained from UK Biobank (Figure \ref{fig:arch_fig}). Given a patient's smoothened Flow-Volume curve $X \in \mathbb{R}^T$ with \ $T$ steps, their demographic information $f_d$, and the correponding target label $y$, we first patchify the input curve into $N$ patches, each of length $P$: $X \in \mathbb{R}^{T} \rightarrow \{p_1, p_2, \ldots, p_N\}, \quad p_i \in \mathbb{R}^{P}, \quad N = \frac{T}{P}$. Each patch $p_i$, which is standardized across all patients, is then passed through a linear projection layer, transforming it into an embedding space $Z_i \in \mathbb{R}^{d_{\text{embed}}}$, where $d_{\text{embed}}$ is the embedding dimension: 
\begin{align*}
\centering
    Z_i &= W p_i + b, \quad W \in \mathbb{R}^{d_{\text{embed}} \times P}, \quad b \in \mathbb{R}^{d_{\text{embed}}}
\end{align*}

Next, following the BERT-style training paradigm \cite{devlin2018bert}, we prepend a learnable classification token, CLS $\in \mathbb{R} ^ {1 \times d_{\text{embed}}}$ to the sequence of patch embeddings and then add positional encodings, $PE_i \in \mathbb{R}^{d_{\text{embed}}}$, to all the tokens:
\begin{align*}
\centering
Z' &= [\text{CLS}+PE_0; Z_1+PE_1; Z_2+PE_2; \ldots; Z_N+PE_N], \quad Z' \in \mathbb{R}^{(N+1) \times d_{\text{embed}}}
\end{align*}
This augmented sequence is then fed into a Transformer encoder layer \cite{vaswani2017attention}, which captures the temporal dependencies within the data, producing the output embeddings $H$.
\begin{align*}
\centering
H = \text{Transformer}(Z', M), H \in \mathbb{R}^{(N+1) \times d_{\text{embed}}}, M \in \mathbb{R}^{(N+1)}
\end{align*}

Here, $M$ represents a mask indicating which of the input patches are padding tokens. Following that, we extract the encoder embedding corresponding to the CLS token from $H$ and pass it through a Multi-Layer perceptron to get the initial assessment: $f_\text{initial} = MLP(H[0])$. We then perform a feature fusion step in which the \revision{CLS token embedding} is concatenated with demographic information. Finally, the fused features are processed using a CatBoost classifier to predict the target probability: $\hat{y} = \text{CatBoost}(f_\text{initial} \oplus f_d)$. CatBoost \cite{prokhorenkova2018catboost} is a gradient boosted decision tree framework that can handle categorical features. Details about the training \revision{and model selection} are in Appendix \ref{training_details}. 

There are multiple baselines that we compare against: (i) \textbf{FEV$_1$/FVC ratio}, which is the summary metric traditionally used as a measure of COPD risk (ii) \textbf{DeepSpiro} \cite{mei2024deep}, which is the prior method based on a Bi-LSTM and a temporal attention layer along with feature fusion (iii) \textbf{MLP (demographic)}, which is a MLP classifier with two hidden layers trained on the demographic features age, sex, smoking status, height (iv) \textbf{MLP (summary stats)}, which is a MLP classifier with two hidden layres trained on the summary statstics derived from the spirometry curve, including FEV$_1$, FVC, and the blow ratio FEV$_1$/FVC.

\vspace{-1.2em}
\section{Results}
The evaluation metric we use is the area under the receiver operating characteristic curve (ROC-AUC). This metric is suitable for our tasks since it handles any imbalance that may be present in the dataset. Figure \ref{fig:auroc_perf} shows the performance of the time-series transformer compared against the baselines along with the standard deviations across five trials. We observe that the time-series transformer outperforms the baselines across all three endpoints. The improvement compared to the strongest baseline, DeepSpiro, is marginal but the time-series transformer allows for greater parallelism during training, as the operations can be batched efficiently. As a result, it is more computationally efficient, with each epoch taking  $\sim$40 seconds, compared to $\sim$180 seconds per epoch for training DeepSpiro. Results for another evaluation metric, the Brier score, are provided in Appendix \ref{brier_score_results}.

\begin{table}[h!]
        \centering
        \begin{tabular}{l|r|r|r}
\hline
                        & COPD risk              & Mortality         & Exacerbation            \\ \hline
FEV$_1$ /FVC Ratio          & 0.759 ± 0.001      & 0.863 ± 0.0204        & 0.805± 0.0075           \\ \hline
MLP (Summary Stats)     & 0.779± 0.006         &   0.909 ± 0.0188         & 0.822± 0.0114           \\ \hline
MLP (Demographic)       & 0.743± 0.005       & 0.817± 0.0133            & 0.801± 0.0076           \\ \hline
DeepSpiro               & 0.817± 0.002       & 0.923± 0.0106              & 0.844± 0.0047           \\ \hline
Time-Series Transformer (Ours) & \textbf{0.828± 0.004} & \textbf{0.929 ± 0.0147} & \textbf{0.851 ± 0.0027} \\ \hline
\end{tabular}
    \caption{Performance of the time-series transformer compared to the other baselines in terms of the area under the ROC curves (ROC-AUC score) across all endpoints. The uncertainty estimates indicate the standard deviation across five independent trials.}
\label{fig:auroc_perf}
\end{table}

\begin{figure}[h!]
\centering
\includegraphics[width=\textwidth]{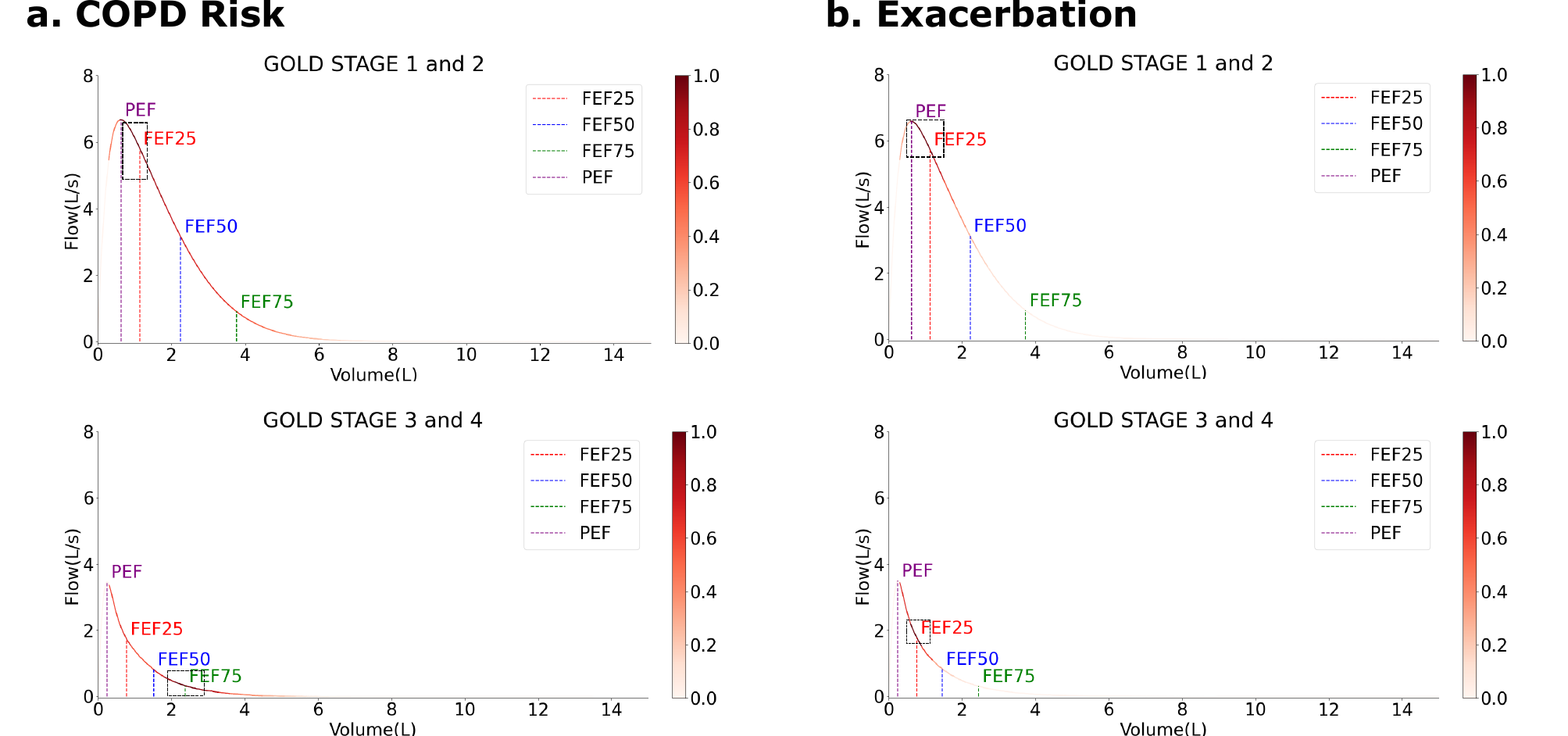}
\caption{Overlaying the attention weights from the transformer encoder onto the Flow-Volume curve to visualize the importance of each patch for the COPD Risk and Exacerbation prediction tasks. The black rectangle in each curve represents the most important patch.}
\label{fig:attn_maps}
\end{figure}

To understand the importance of each patch to the model predictions, we consider the mean attention weights of the CLS token across all the encoder layers with each of the other patch tokens after normalizing them using a softmax function. We then overlay these attention scores on top of the original Flow-Volume curves to visualize the importance of each patch. To better understand the model predictions for different patient cohorts, we stratify the patients based on the severity of COPD using the  Global Initiative for Chronic Obstructive Lung Disease (GOLD) criteria. We consider two cohorts: (i) GOLD Stage 1 and 2; Mild to Moderate COPD (FEV$_1$ \% predicted >= 50) and (ii) GOLD Stage 3 and 4; Moderate to Severe COPD (FEV$_1$\% predicted < 50). For each cohort, the attention scores are averaged across all patients. We also annotate the curve with some important markers, like the PEF and the FEF25/50/75 which represents Forced expiratory flow \% exhaled (Figure \ref{fig:attn_maps}). The black boxes represent the part of the curve that is the most important for that specific prediction task (due to space constraints, results for the mortality endpoint are deferred to Appendix \ref{mortality_results}). For predicting COPD risk, different segments of the spirometry curve hold varying levels of importance depending on the GOLD stage. For GOLD stages 1 and 2, the initial segment of the curve, between the PEF and FEF25, is particularly significant, whereas for GOLD stages 3 and 4, the latter portion of the curve, between FEF50 and FEF75, becomes more relevant. When predicting exacerbation, the early segment (PEF to FEF25) is crucial for GOLD stages 1 and 2, while the middle segment (FEF25 to FEF50) is more pertinent for stages 3 and 4.


\vspace{-1.2em}
\section{Clinical Insights}
After identifying the significance of various curve segments for model predictions, we consulted a board-certified pulmonologist to gain clinical insights into the results. The initial portion of the flow-volume curve reflects airflow from the larger airways (e.g., main bronchi), while the tail end of the curve corresponds to airflow from the smaller airways (e.g., alveoli, bronchioles). Key features of the curve that are characteristic of COPD include: (i) a steep initial decline in flow at high lung volumes (following the PEF), (ii) a scooped or concave shape indicative of airflow limitation, (iii) significant displacement in the middle portion of the curve, and (iv) a flattening toward the end (between FEF50 and FEF75), indicative of reduced expiratory flow rate at low lung volumes, which may indicate respiratory disorders. Our findings reveal that the model prioritizes segments occurring after the PEF, with none of the pre-PEF segments identified as significant, suggesting the model is correctly focusing on curve aspects that are clinically relevant for COPD diagnosis. While further evaluation of the model's explanations is necessary, the current observations align with established clinical knowledge of COPD-related outcomes.

\section{Conclusion}
In this study, we investigate the advantages of using raw spirometry curves over traditional summary measures for predicting COPD-related outcomes, employing a transformer-based deep learning framework. Our model outperforms previous methods while demonstrating greater computational efficiency. By leveraging the attention weights of the transformer encoder, we incorporate interpretability into the framework, enabling us to identify critical regions of the curve that influence model predictions. The identified regions correspond with clinical insights provided by a board-certified pulmonologist, highlighting the model's alignment with established medical understanding. This work establishes a foundation for a more comprehensive exploration of therapeutic biomarkers for COPD.
\clearpage

\bibliographystyle{plain}

\bibliography{main}
\appendix

\clearpage
\section{Appendix}
\subsection{Related Work}
Since COPD is a high-prevalence disease with multiple modeling challenges, there has been prior research done on detecting COPD using deep learning techniques on different types of data modalities. One study employed a recursive feature elimination technique evaluated with several classification models to distinguish early-stage COPD patients from those in advanced stages using clinical and demographic information \cite{hussain2021detection}. Another one  conducted an empirical pulmonolgy study of a representative sample of 132 patients for the prediction of COPD and asthma using a random forest classifier \cite{spathis2019diagnosing}. The study utilized 22 different features about the patients, ranging from demographic and medical information to special lung summary measurements obtained from a spirometer. One method aimed at predicting the risk factors for COPD diagnosis using machine learning techniques like gradient boosted decision trees and logisitic regression models \cite{muro2021machine}. The dataset consisted of the annual medical check-up of Hitachi employees during a 20 year period. One other method demonstrated the potential of using a single commercial metal oxide gas sensor, TGS-2602STMS, to capture comprehensive breath information for COPD detection by modeling it using a Support Vector Machine (SVM) classifier \cite{mahdavi2024single}. There is a prior work that trained, validated and compared three different type of networks: a radial basis function (RBF), a k-means classifier, and a probabilistic neural network (PNN) to predict exacerbations of COPD from daily
responses to a health questionnaire spanning 6 months \cite{fernandez2015machine}. Focusing on imaging modalities, one study developed a method to detect and quantify emphysema in COPD patients using x-ray dark-field images, CT images, and pulmonary function test measurements \cite{willer2021x}. Another imaging study developed a deep learning method to stage COPD severity by quantifying emphysema and air-trapping in CT images using a co-registration and lung segmentation algorithm \cite{hasenstab2021automated}. However, none of these prior methods focused on utilizing the raw spirometry blow values for COPD prediction.

There have been prior works that have tried modeling the raw spirogram using deep learning techniques. One method used a CNN-based ResNet-1D model to predict COPD liability scores without requiring specific domain knowledge \cite{cosentino2023inference}. We use the same dataset and labeling procedures used in this study. However, this work does not perform any explainability analysis to understand the reasoning processes of the trained classifier. Another method, called DeepSpiro, built on the ResNet-1D model by introducing several modifications like Bi-LSTM for modeling the long range dependencies, a temporal attention layer for providing interpretability, and incorporating demographic information \cite{mei2024deep}. This is the main method that we build on and compare against.

\subsection{Data Preprocessing}
\label{preprocess}
From the initial Volume-Time measurements, we employ a number of preprocessing steps to curate the input to our model. To ensure that the blows are accurately measured, we consult UKB field 3061 which indicates the acceptability of each blow result. If the value in this field is either 0 or 32, the blow is considered to be valid. To control the quality of the measurements, we look at the FEV$_1$, FVC, and PEF values derived from the Time-Volume curve and drop any blow which is in the extreme tail of all observed values (top or bottom 0.5\%). After completing these preprocessing steps, we get 356,848 patients with valid spirograms. 

To process the Time-Volume curve, we first convert the measurements from milliliters to liters and apply a 1D-Gaussian smoothing with a standard deviation of 1 to enhance the stability of the curves. Following this, based on respiratory physiology and prior works \cite{miller2005standardisation, barreiro2004approach, mei2024deep, cosentino2023inference}, we convert the Volume-Time to Flow-Time curves by taking the first derivative with respect to time using finite differences. Specificially, if $V(t)$ represents the Volume measurements at time $t$ and $\Delta t$ represents the time interval, the flow $F(t) = \frac{V(t+\delta t) - V(t)}{\delta t}$. The flow data is then linearly interpolated to ensure consistency between the Flow-Time and Volume-Time curves. The resulting volume-time and flow-time curves are then combined to generate a one-dimensional flow-volume curve which is the input to our model (Figure \ref{fig:arch_fig}). To ensure that all the curves are of the same length, we right-pad the shorter curves by zero upto the length of the longest curve in the dataset. We also extract demographic information like age, sex, smoking status, and height, which are all collected at the same time as the date of spirometry and have been shown previously to be predictive of COPD. \revision{The prevalence of COPD has been shown to be higher in older people and females \cite{levin2020prevalence, perez2020sex}. Smoking status is a known confounder for COPD \cite{godtfredsen2008copd, nielsen2024copd} so we adjust for it by including it as a feature in our model. Patients with different heights will have different lung capacities which can affect the COPD status so it is also included as a demographic features. During modeling, age and height are represented as integers while smoking status and sex are represented as binary variables.}

\subsection{Details of Label Generation}
\label{label_details}
To obtain the labels for training, we use the medical records of the patients present in the UK Biobank dataset, specifically the self report records, the hospital inpatient records, and the primary care records. The self report records (UKB field 20002) contain non-cancerous illness codes which are placed in a coding tree, either by the participant or a trained nurse. Presence of codes 1112, 1113, or 1472 indicates a positive COPD-status. For the hospital inpatient records (fields 41270 and 41271) we consider the presence of COPD-related ICD-9 and ICD-10 codes. COPD status from GP clinical records (field 42040) is obtained after the read v2 and read v3 codes are mapped to the corresponding ICD-10 codes using the TRUD mappings (data codings 1835 and 1834). The exacerbation label only includes cases with COPD as primary cause of hospitalization after the spirometry test date. COPD-related Mortality information is obtained from UKB field 40000. 

Following the filtering codes used in \cite{cosentino2023inference}, the ICD-10 codes considered as being related to COPD include: J43 (Emphysema), J440 (Chronic obstructive pulmonary disease with acute lower respiratory infection), J441 (Chronic obstructive pulmonary disease with (acute) exacerbation), J449 (Chronic obstructive pulmonary disease, unspecified). The ICD-9 codes considered include: 492 (Emphysema), 496 (Chronic Airway Obstruction). For the mortality endpoint, ICD-10 code J41 (Chronic Bronchitis) is also considered.

\subsection{Model Selection and Training Details}
\label{training_details}
\revision{For the disease classifier after the feature fusion step, we decided to use the CatBoost framework because of its ability to handle categorical features natively. Other boosting algorithms like XGBoost \cite{chen2016xgboost} and LightGBM \cite{ke2017lightgbm} require manually encoding of categorical features (e.g. one-hot encoding). CatBoost is able to encode categorical features without causing data leakage, resulting in reduced overfitting. Also, CatBoost has been shown to be robust to hyperparameter tuning \cite{prokhorenkova2018catboost}.}

For training, after doing a hyperparamater sweep, we use a patch length $P=30$ and an embedding dimension $d_\text{embed} = 200$ with 2 transformer encoder layers, each with 2 multiattention heads. For the CatBoost classifier, we use the default parameter values provided as part of the library. The standard cross-entropy loss is used to train the model for 30 epochs with the Adam optimizer \cite{kingma2014adam} and a learning rate of $10^{-5}$, using a train/test/val split of 80/10/10. \revision{During the feature fusion step, we tested with using both the normalized and the raw values for the demographic features to concatenate with the transformer's output embedding and got similar performance in both scenarios.}

\subsection{Additional Evaluation Metrics}
\label{brier_score_results}
In addition to the ROC-AUC scores, we also use the Brier score as an evaluation metric. The Brier score is a metric used to assess the accuracy of probabilistic predictions. It measures the mean squared difference between predicted probabilities and the actual outcomes (typically binary, where 0 or 1 represents the observed result). A lower Brier score indicates that the model is better calibrated and has more accurate predictions, with a score of 0 representing perfect predictions.

\[
\text{Brier Score} = \frac{1}{N} \sum_{i=1}^{N} (\hat{p}_i - y_i)^2
\]

where  $\hat{p}_i$ is the predicted probability and $y_i$ is the target label for the \( i \)-th instance. We observe that our method is able to outperform the the other baselines across all endpoints except for the mortality, but the brier score is  still close to zero, indicating that the model is confident in its predictions (Supplementary Table \ref{table:brier_score}).

\begin{supptable}[h!]
        \centering
        \begin{tabular}{l|r|r|r}
\hline
                        & COPD risk           & Mortality               & Exacerbation            \\ \hline
FEV$_1$ /FVC Ratio          & 0.0776 ± 0.0001         & 0.065 ± 0.0000          & 0.0724± 0.0001           \\ \hline
MLP (Summary Stats)     & 0.0335± 0.0000          & 0.0012 ± 0.0000          & 0.0209± 0.0000           \\ \hline
MLP (Demographic)       & 0.0317± 0.0001          & \textbf{0.0011± 0.0000}           & 0.0203 ± 0.0002           \\ \hline
DeepSpiro               & 0.0281± 0.0002          & 0.0061± 0.0001           & 0.0202± 0.0001           \\ \hline
Time-Series Transformer (Ours) & \textbf{0.0280± 0.0002} & 0.0045 ± 0.0001 & \textbf{0.0199 ± 0.0000} \\ \hline
\end{tabular}
    \caption{Performance of the time-series transformer compared to the other baselines in terms of the brier score across all endpoints. The uncertainty estimates indicate the standard deviations across five independent trials.}
\label{table:brier_score}
\end{supptable}

\subsection{Results for Mortality endpoint}
\label{mortality_results}
Here we present the interpretability results for the mortality endpoint. In the attention maps for predicting mortality, the middle portion of the curve (FEF25 to FEF50) is key for GOLD stages 1 and 2, whereas the final portion (FEF50 to FEF75) holds greater importance for stages 3 and 4 (Supplementary Figure \ref{fig:attn_maps_mortality}). This still aligns with the underlying medical knowledge as observed for the COPD risk and exacerbation endpoints.


\begin{suppfigure}[h!]
\centering
\includegraphics[width=0.8\textwidth]{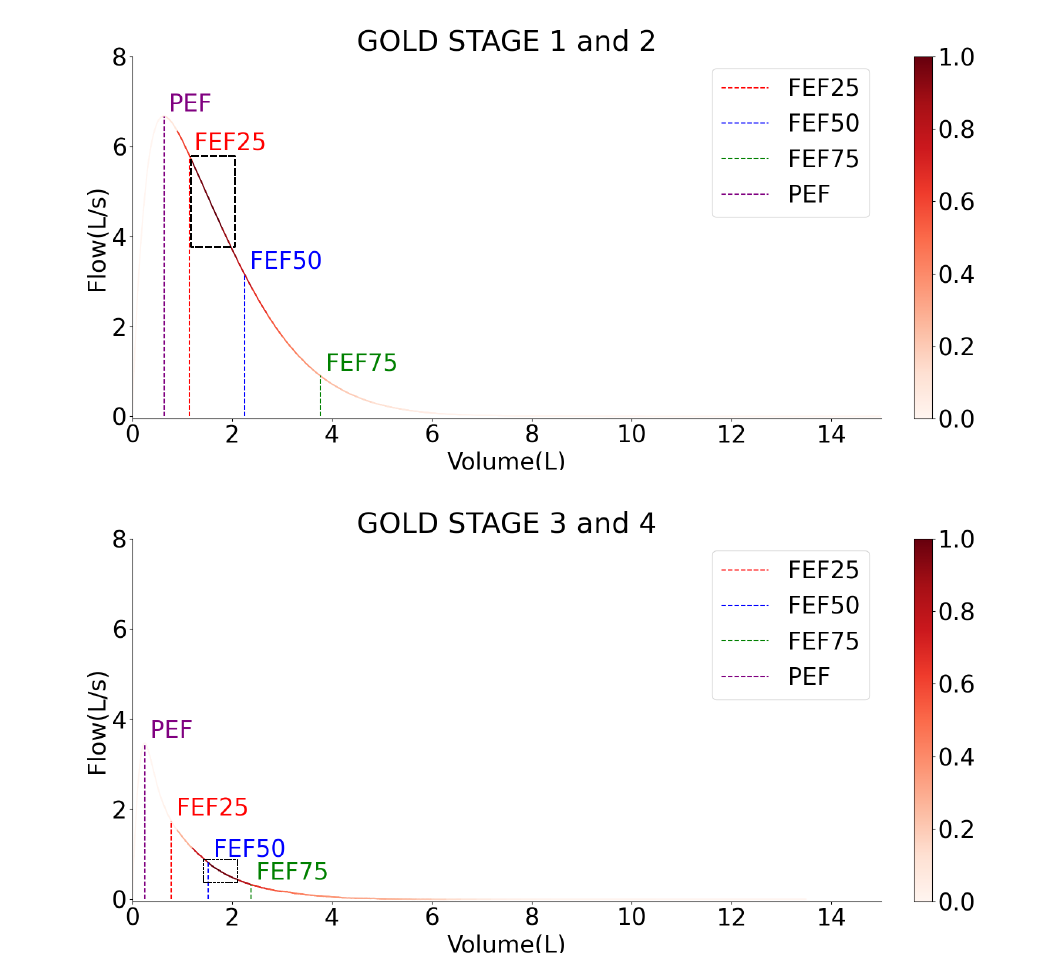}
\caption{Overlaying the attention weights from the transformer encoder onto the Flow-Volume curve for the Mortality endpoint. The black rectangle in each curve represents the most important patch.}
\label{fig:attn_maps_mortality}
\end{suppfigure}
\end{document}